\documentclass[letterpaper, 10 pt, conference]{ieeeconf}  

\IEEEoverridecommandlockouts                              

\usepackage{graphics} 
\usepackage{epsfig} 
\usepackage{times} 
\usepackage{amsmath} 
\usepackage{amssymb}  
\usepackage{bm}
\usepackage{color}
\usepackage{cite}
\usepackage[linesnumbered,ruled]{algorithm2e}
\usepackage{ulem} 
\usepackage[colorlinks=true,linkcolor=black,citecolor=blue,urlcolor=blue,]{hyperref}
\usepackage[doipre={doi:~}]{uri}
\usepackage{float}
\usepackage{booktabs}
\usepackage{multirow}
\usepackage{mathrsfs}
\usepackage[utf8]{inputenc}
\usepackage[caption=false, font=footnotesize]{subfig}
\usepackage{graphicx}
\usepackage{pifont}

\newcounter{RNum}
\renewcommand{\theRNum}{\arabic{RNum}}
\newcommand{\Remark}{\noindent\textit{\textbf{Remark}~\refstepcounter{RNum}\textbf{\theRNum}: }}
\usepackage{soul}
\newcommand{\NoOne}[1]{\textcolor{red}{#1}}
\newcommand{\NoTwo}[1]{\textcolor{green}{#1}}
\newcommand{\NoThree}[1]{\textcolor{blue}{#1}}
 
\soulregister\NoOne7
\soulregister\NoTwo7
\soulregister\NoThree7
\soulregister\Remark7
\usepackage{flushend}

\title{\LARGE \bf
HighlightNet: Highlighting Low-Light Potential Features \\ for Real-Time UAV Tracking
}

\author{Changhong Fu$^{*}$, Haolin Dong, Junjie Ye, Guangze Zheng, Sihang Li, and Jilin Zhao
\thanks{$^{*}$Corresponding author}
\thanks{The authors are with the School of Mechanical Engineering, Tongji University, Shanghai 201804, China.
        {\tt\small changhongfu@tongji.edu.cn}}%
}

\begin{document}

\maketitle
\thispagestyle{empty}
\pagestyle{empty}

\begin{abstract}
Low-light environments have posed a formidable challenge for robust unmanned aerial vehicle (UAV) tracking even with state-of-the-art (SOTA) trackers since the potential image features are hard to extract under adverse light conditions. Besides, due to the low visibility, accurate online selection of the object also becomes extremely difficult for human monitors to initialize UAV tracking in ground control stations. To solve these problems, this work proposes a novel enhancer, \textit{i.e.}, HighlightNet, to light up potential objects for both human operators and UAV trackers. By employing Transformer, HighlightNet can adjust enhancement parameters according to global features and is thus adaptive for the illumination variation. Pixel-level range mask is introduced to make HighlightNet more focused on the enhancement of the tracking object and regions without light sources. Furthermore, a soft truncation mechanism is built to prevent background noise from being mistaken for crucial features. Evaluations on image enhancement benchmarks demonstrate HighlightNet has advantages in facilitating human perception. Experiments on the public UAVDark135 benchmark show that HightlightNet is more suitable for UAV tracking tasks than other state-of-the-art (SOTA) low-light enhancers. In addition, real-world tests on a typical UAV platform verify HightlightNet's practicability and efficiency in nighttime aerial tracking-related applications. The code and demo videos are available at \url{https://github.com/vision4robotics/HighlightNet}.
\end{abstract}
\section{Introduction} \label{sec:intro}
\begin{figure}[!t]	
\centering
\includegraphics[width=0.98\linewidth]{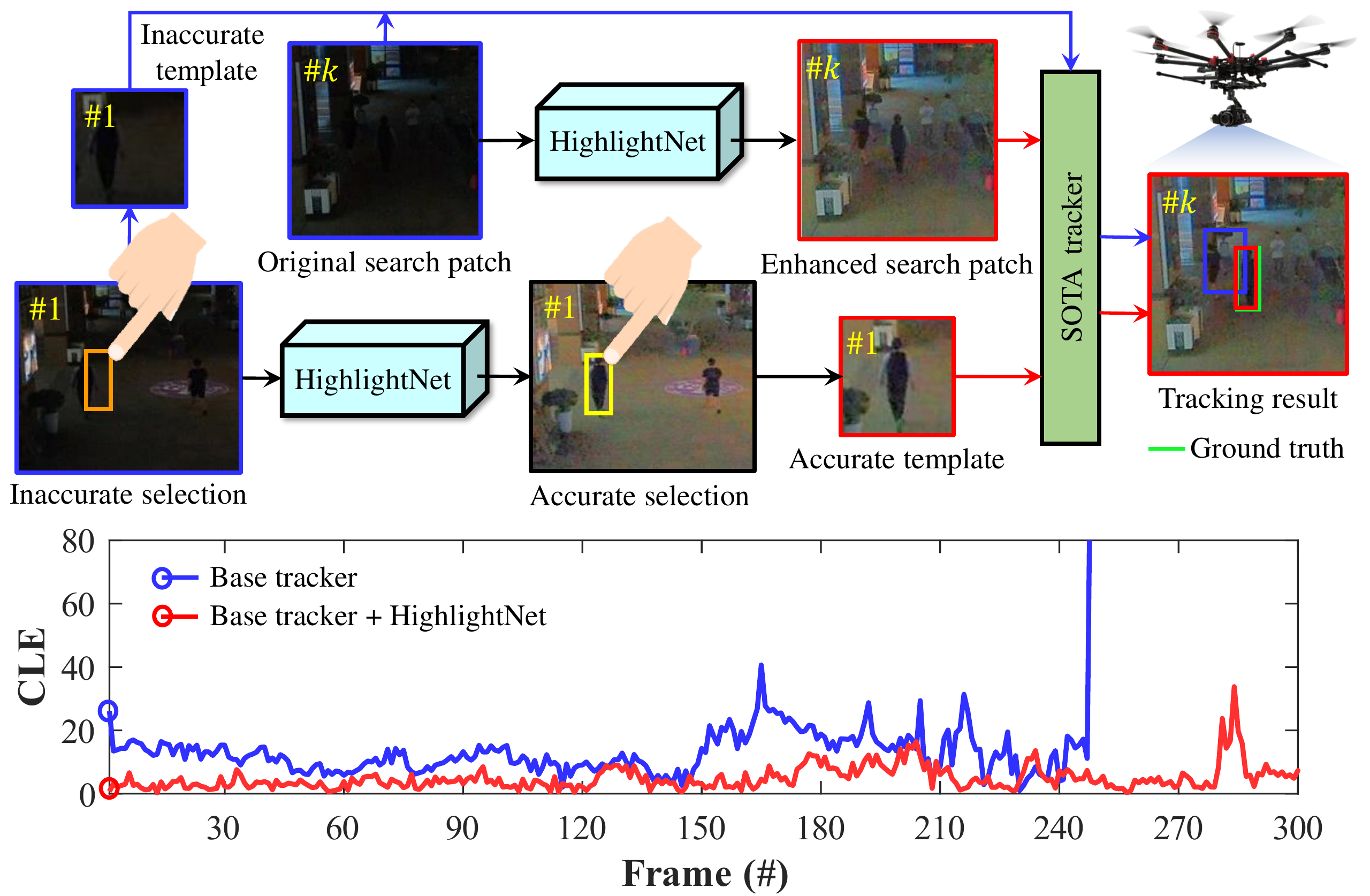}
\setlength{\abovecaptionskip}{-12pt} 
\caption
{
	Comparison of the overall tracking results from the baseline tracker with HighlightNet (in \textcolor{red}{red}) or not (in \textcolor{blue}{blue}). Operators may make an inaccurate selection due to the low illumination. With the help of HighlightNet, human operators can conduct an accurate selection of targets, and the SOTA tracker can achieve promising results, as also shown in center location error (CLE) curves.  
}
\vspace{-12pt}
\label{fig:fig1}

\end{figure}

Object tracking has been widely used in a variety of sectors, most representative in UAV applications for target following~\cite{Cheng2017IROS}, autonomous landing~\cite{LAND}, and self-localization~\cite{Ye_2021_TIE}. SOTA tracking approaches~\cite{Li_2018_CVPR, Cao2021ICCV} have achieved outstanding performance under favorable illumination. However, the degradation of these trackers' performance in low-light conditions has been discussed in recent studies~\cite{li2021allday,Ye2021IROS}. On the one hand, as shown in Fig.~\ref{fig:fig1}, human operators can hardly make an accurate initial annotation for trackers. On the other hand, due to the quick mobility of both the UAV and the tracked object, as well as numerous challenges such as occlusion, noise, illumination variation, robust and precise tracking in nighttime conditions has remained a difficult task. An effective approach to light up the target object for both human operators and trackers is in dire need.


A low-light enhancer can be a solution to this problem. However, enhancement of objects in dark areas may lead to overexposure problems in regions under favorable illumination, which may cause the destruction of potential features. Besides, artificial light sources are omnipresent in low-light conditions, which divided the image into regions with different illumination. In UAV object tracking, this problem is especially crucial because of UAVs' wide vision field. When the target enters over-exposed or extreme dark regions, it is more difficult for the tracker to recognize the object. In this case, this work proposes a range mask to distinguish the enhancement of regions with or without artificial light sources. Moreover, experiments on typical nighttime UAV scenes show that the enhancer can focus on the enhancement of tracking objects according to the mask.

Only highlighting tracking objects and other effective local features is not enough. In nighttime UAV applications, it is inevitable that noises will damage the structural details of images and even be mistaken for tracking objects. Moreover, with excessive background information in UAVs' wide vision field, more background noises are also introduced. However, existing low-light enhancers can hardly distinguish the noise from effective features. To avoid over-enhancement of noise, we consequently design a dark area truncation function, which produces an anti-noise mask to filter unwanted noise.  

Apart from effective local features, global information is also essential for low-light enhancers. Due to the high mobility of UAVs, rapid changes of scenarios can hardly be avoided, which may cause severe global illumination variation. Nevertheless, most low-light enhancers~\cite{LIME,RetinexNet,Zhang2019ACM,EnlightenGAN,Guo_2020_CVPR} are designed for stable photography scenes. Without adjustment mechanisms, their effectiveness is unsatisfying while dealing with illumination variation. To address this problem, Transformer~\cite{vaswani2017nips} is introduced to dynamically regulate parameters according to global illumination information.




Another difficulty of UAV image enhancement is the limitation of computational resources. In consideration of the take-off weight, UAVs can hardly carry a high-power computing platform. Approaches with a high computation complexity have difficulty being processed in real-time on UAVs. To improve the computing efficiency, the main workflow of HighlightNet is based on the gray channel rather than the RGB channels. In addition, for supervised training, most low-light enhancers use paired data. In order to reduce the high expense of gathering enough paired data for UAV conditions, HighlightNet is designed to be trained with unpaired data.

The contributions of this work are summarized as follows:
\begin{itemize}
	\item An adaptive enhancer HighlightNet is constructed to facilitate both online object selection and UAV tracking in low-light conditions. 
	\item The range mask and the anti-noise mask are designed to highlight the object and filter unwanted noise.
	\item A dynamic parameter adjustment method based on an efficient Transformer is proposed to process global features for illumination retouching.
	\item Comprehensive evaluation on image enhancement test sets, UAV nighttime tracking benchmarks, and real-world experiments demonstrate that HighlightNet has advantages in facilitating human operator perception and improving trackers' nighttime performance.
\end{itemize}

\section{Related Works}

\subsection{UAV Tracking}
Object tracking methods include correlation filter-based approaches~\cite{Danelljan2017CVPR,zheng2021ICRA} and convolutional neural network (CNN)-based approaches~\cite{LiBo_2019_CVPR, Xu_2020_AAAI, Cao2021ICCV}. Among them, due to the promising tracking performance, methods based on Siamese networks have been widely used in UAV tracking. SiamAPN~\cite{Fu_2021_TGRS} is a Siamese-based no-prior two-stage method for adaptive anchor proposing. \cite{Li_2018_CVPR,LiBo_2019_CVPR,Ren2017tpami} employ the region proposal network (RPN) as anchor-based trackers to further improve tracking accuracy. HiFT~\cite{Cao2021ICCV} introduces a feature Transformer network to obtain hierarchical similarity maps. Ad$^2$Attack~\cite{AD2} proposes a novel adaptive adversarial attack approach to help increase awareness of the potential risk. TCTrack~\cite{TCT} designs a comprehensive framework to fully utilize temporal contexts for UAV tracking. In spite of their high accuracy in daytime conditions, their performance is unsatisfying in low-light conditions. Damage to potential features in nighttime conditions makes their robustness drop greatly. What's worse, precise online target selection is nearly impossible in extreme low-light conditions. Therefore, image enhancement technology before tracking is urgently needed.

\subsection{Low-light Image Enhancement}
The theory of Retinex~\cite{Edwin_1977_retinex} has evolved into a variety of low-light enhancement approaches.  LIME~\cite{LIME} estimates a coarse illumination map by extracting the maximum of each pixel in RGB channels, which is subsequently refined by a structure prior. Based on Retinex, convolutional neural networks (CNNs) are introduced and remarkably improve low-light performance~\cite{RetinexNet, Zhang2019ACM}. The first CNN model LL-Net \cite{lore2017llnet} employs an autoencoder to learn denoising and light enhancement simultaneously. DeepUPE \cite{DUPE} introduces intermediate illumination to connect the input and the anticipated enhancement result. The method in~\cite{LEARN} takes noise into consideration and realizes satisfying performance in enhancing practical noisy low-light images. However, because of the internal locality of convolution operations, CNNs are not suitable for processing global features.

Another disadvantage of most CNN-based methods is the requirement of paired data for supervised training. In consideration of the high investment caused by collecting sufficient paired data for UAV conditions, low-light enhancement by paired data is impractical. Therefore, advanced methods turn to unsupervised training. EnlightenGAN~\cite{EnlightenGAN} employs a generative adversarial network (GAN) so it can be trained using adversarial loss as an attention-based U-Net. Zero-DCE~\cite{Guo_2020_CVPR} employs image-specific curve estimation and has high efficiency, also trained with unpaired data. RUAS~\cite{RUAS} can also be trained with unpaired data by employing a cooperative reference-free learning strategy. Nevertheless, in low-light conditions, the performance of SOTA enhancers in target object selection and target tracking can hardly meet our satisfaction. This can be partly attributed to the neglect of potential effective features.

\subsection{Low-light UAV Tracking}

Though UAV tracking tasks in nighttime scenes are crucial, only several approaches~\cite{Ye2021IROS, SCT, li2021allday,Ye2022CVPR} to improve UAV tracking performance in low-light conditions have been proposed. ADTrack~\cite{li2021allday} serves for correlation filter (CF) trackers, while Darklighter~\cite{Ye2021IROS} and SCT~\cite{SCT} are mainly designed for trackers based on Siamese network. Darklighter~\cite{Ye2021IROS} constructs a lightweight map estimation network to cope with poor illumination and noise. Applying task-tailored training and a Transformer structure, SCT~\cite{SCT} realizes stable nighttime tracking. The basic idea of these three methods is to employ a plug-and-play enhancer as a preprocessing step for UAV trackers. UDAT~\cite{Ye2022CVPR} instead considers the nighttime tracking problem as an unsupervised domain adaptation task. Despite the progress, existing approaches generally ignore the improvement of human perception, which plays a vital role in the step of online object selection. Moreover, without adjustment mechanisms and denoising modules, these methods can hardly deal with illumination variation and background noises.

\begin{figure*}[!t]	
	\includegraphics[width=0.98\linewidth]{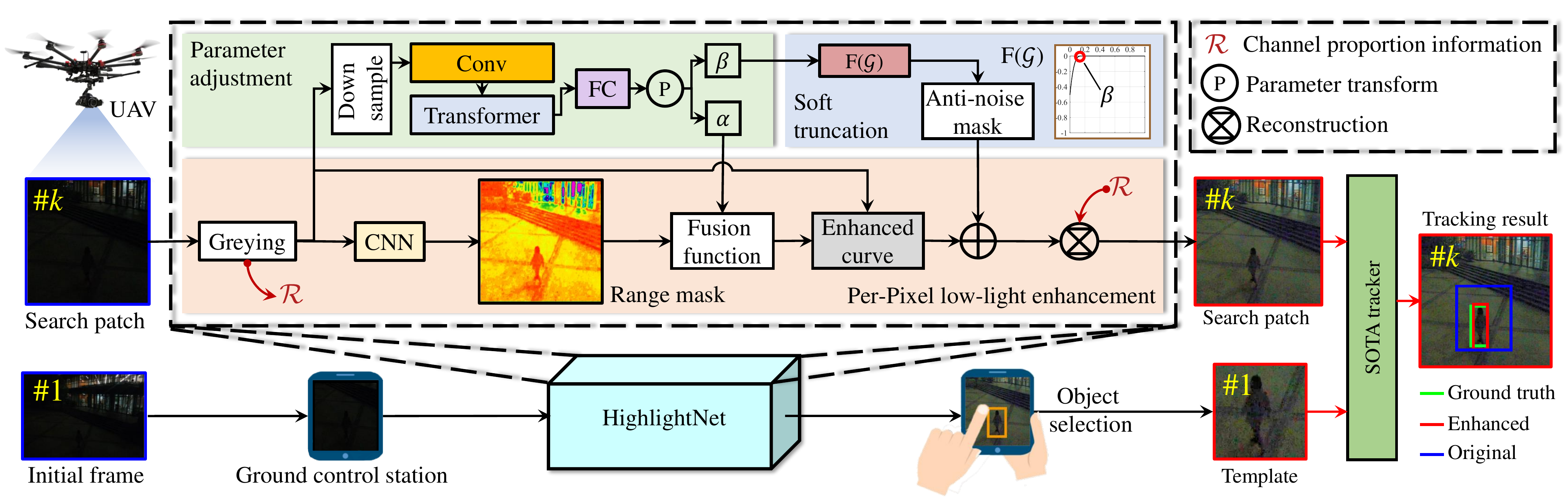}
	\setlength{\abovecaptionskip}{-2pt} 
	\caption
	{
		Overview of our HighlightNet pipeline. Given a low-light image patch, HighlightNet includes three novel modules, \textit{i.e.}, per-pixel low-light enhancement, Transform-based parameter adjustment, and soft truncation, to highlight potential features for both human operators and trackers with slight computational consumption. The bottom row serves for template selection by humans.
	}
	\label{fig:main}
\end{figure*}

\section{Methodology}
This work proposes a Transformer-based adaptive enhancer called HighlightNet to highlight potential features for low-light conditions. As shown in Fig.~\ref{fig:main}, to serve both human operators and UAV trackers, the template patch and search patch are processed separately on ground control stations and UAVs. The proposed HighlightNet includes three novel modules. The Transformer-based parameter adjustment module obtains the constraint $\alpha$ and the truncation threshold $\beta$ by processing the downsampled gray-scale image. In the enhancement module, a range mask is constructed by a convolutional neural network (CNN). The mask is fused with constraint $\alpha$ by a fusion function to adapt the enhanced curve. As the mask corresponds to the original image pixel by pixel, each pixel has a unique enhancement range. The truncation threshold $\beta$ acts on the soft truncation module. By the truncation function, the gray-scale image is translated to the anti-noise image. To improve computational efficiency, the main enhancement workflow is based on the gray-scale image. Therefore, the color information is first translated to channel proportion information and stored.

\subsection{Per-Pixel Low-Light Enhancement}
To focus on the enhancement of objects, this module is mainly designed for predicting the accurate enhancement for each pixel. Therefore, the range mask $\mathbf{M}_{i} \in \mathbb{R}^{1 \times H_{i} \times W_{i}}$ is with the same resolution as the input gray-scale image $\mathbf{G}_{i} \in \mathbb{R}^{1 \times H_{i} \times W_{i}}$. We employ a CNN with symmetrical concatenation to obtain the range mask $\mathbf{M}_{i}$. It is constructed by seven convolutional layers. In each layer, the ReLU activation function is followed by 4 convolutional kernels of size $3\times 3$. The sigmoid activation function is then applied to the final layer and produces the range mask $\mathbf{M}_{i}$. The range mask $\mathbf{M}_{i}$ is then fused with constraint $\alpha$ by the fusion function to fit the enhanced curve.

As a pixel-level operation, the enhanced curve has a great influence on the computational efficiency of HighlightNet. Considering the scarce computational resources of UAV platforms, gamma transform is employed as the enhanced curve. It can realize large-scale enhancement without iteration. A standard gamma transform can be expressed as:
\begin{equation}\label{equ:gamma}
	\mathbf{O}_{i}\left ( \textit{r,c} \right )= \mathbf{G}_{i}\left ( \textit{r,c} \right )^{\gamma_{i} \left ( \textit{r,c} \right ) }, \, 0\leqslant \textit{r}< H_{i} \, ,\,  0\leqslant \textit{c}< W_{i}
	\quad,
\end{equation}
where $\mathbf{G}_{i}\left ( \textit{r,c} \right )$ is the value of each pixel in gray-scale image $\mathbf{G}_{i}$, $\gamma \left ( \textit{r,c} \right )$ is the output of the fusion function, $\mathbf{O}_{i}\left ( \textit{r,c} \right )$ is the output of this enhancement module, $H_{i}$ and $W_{i}$ are the height and width of the image. Each pixel of the gray-scale image $\mathbf{G}_{i}$ is normalized to [0,1] before calculation. After designating the enhanced curve, the fusion function can be designed.

The fusion function can change the codomain of each pixel's value in the range mask $\mathbf{M}_{i}$. Because of the sigmoid activation function, the value of pixels in the range mask $\mathbf{M}_{i}$ is limited between 0 and 1. However, for the enhanced curve defined in Eq.~(\ref{equ:gamma}), the input $\gamma \left ( \textit{r,c} \right )$ should be between the constraint $\alpha$ and 1. Therefore, the fusion function can be expressed as: 
\begin{equation}\label{equ:fusion}
	\gamma_{i}\left ( \textit{r,c} \right )= \alpha_{i} ^{\mathbf{M}_{i}\left ( \textit{r,c} \right ) }, \, 0\leqslant \textit{r}< H_{i} \, ,\,  0\leqslant \textit{c}< W_{i}
	\quad,
\end{equation}
where $\mathbf{M}_{i}\left ( \textit{r,c} \right )$ are the value of each pixel in the range mask $\mathbf{M}_{i}$, $\alpha$ is the constraint obtained by the Transformer-based branch, $\gamma \left ( \textit{r,c} \right )$ is the input of the enhanced curve Eq.~(\ref{equ:gamma}). The fusion function and the enhanced curve are not fixed, they can be changed to adapt to different tasks. The reason why we choose gamma transform is mainly for computational efficiency.

\begin{figure}[!b]	
	\centering
	\includegraphics[width=0.98\linewidth]{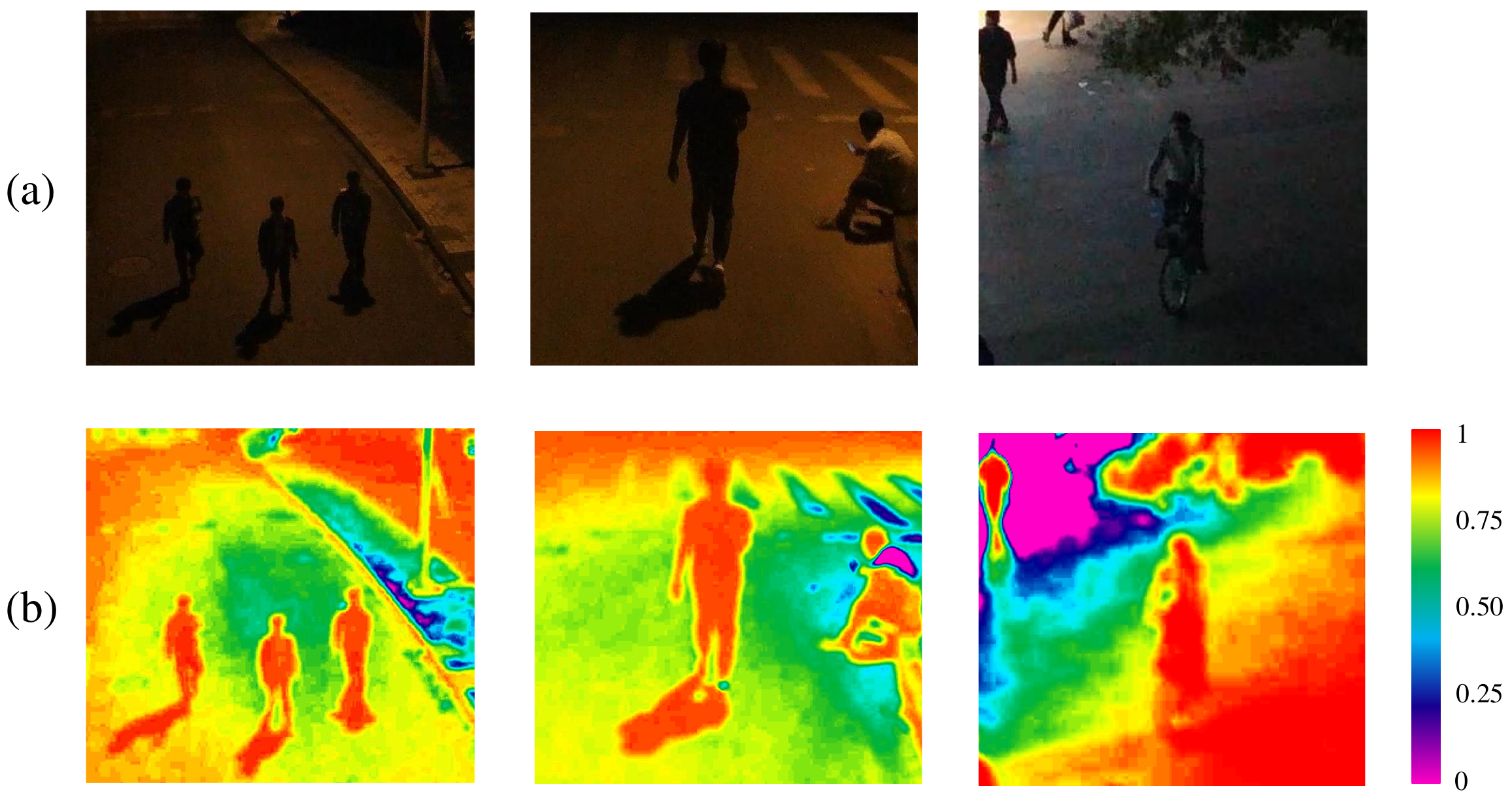}
	\setlength{\abovecaptionskip}{-2pt} 
	
	\caption
	{
		Three examples of the pixel-wise range mask. (a) denotes original image. (b) denotes range mask, represented by heatmaps. For visualization, we select a particular region of the image and normalize the values to the range of [0, 1]. The range mask conducts HighlightNet to focus on the enhancement of tracking objects and regions without light resources.
	}
	\label{fig:mask}
	\vspace{-4pt}
\end{figure}

Three examples of the range mask are presented in Fig.~\ref{fig:mask}. As shown, the range mask makes the enhancer focus on the enhancement of the potential objects. Moreover, the enhanced result exposes the features in shadow and maintains the areas with artificial light sources, which improves the brightness uniformity. Since the shadow is illuminated by HighlightNet, the tracker won't lose the object when it enters the dark areas without light resources.

\begin{figure}[!t]	
	\centering
	\includegraphics[width=0.98\linewidth]{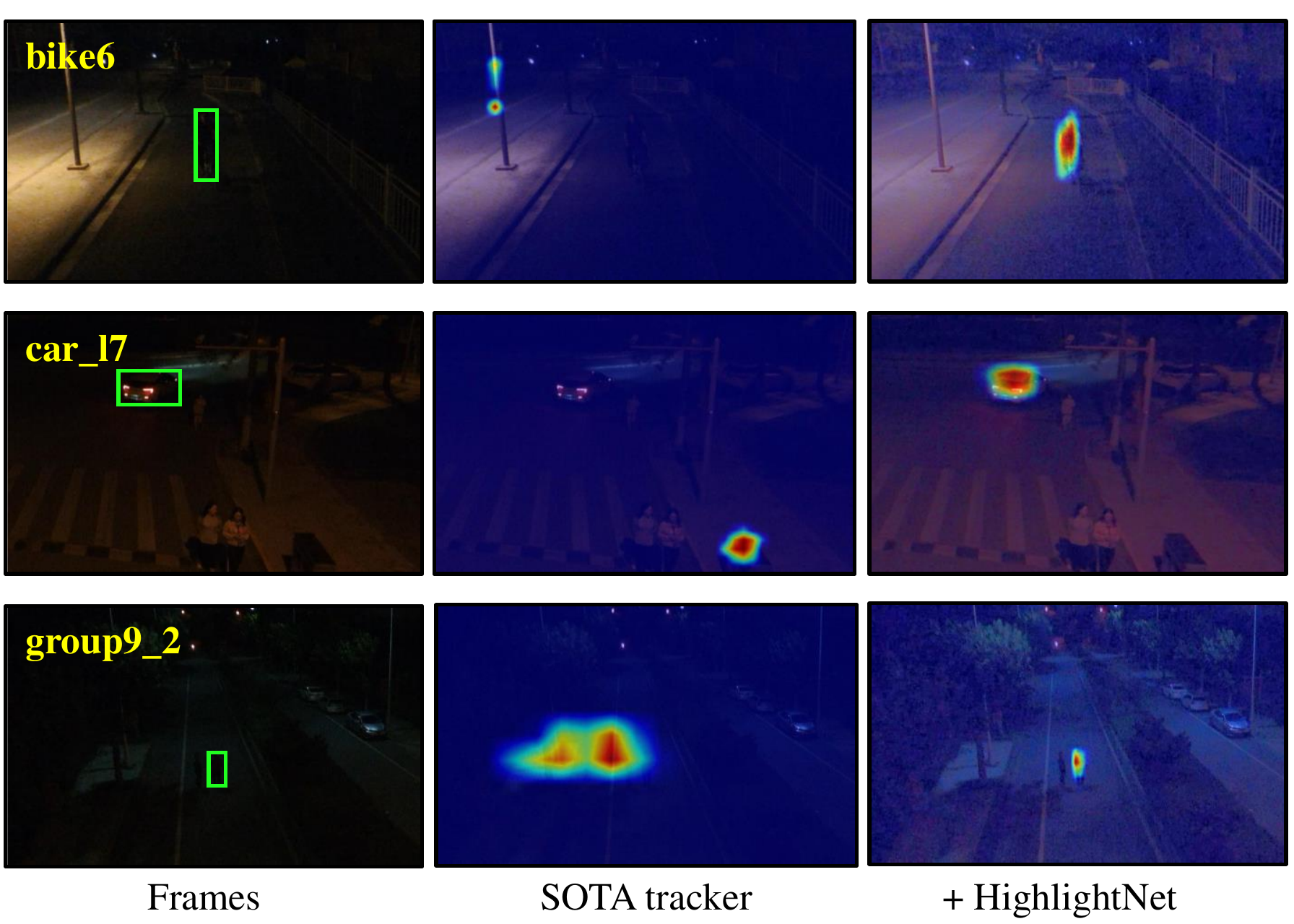}
	\setlength{\abovecaptionskip}{-2.5pt}  
	
	\caption
	{
		Comparison of confidence maps generated by the SOTA tracker with and without HighlightNet. The tracked objects are shown by the \textcolor{green}{green} boxes in the first column. The sequence names are provided in the top left corner of the original frames from the public UAVDark135 benchmark. Without HighlightNet, the base tracker loses its capacity to recognize objects in low-light conditions.
	}
	\label{fig:fig4}
	\vspace{-0.4cm}
\end{figure}

\Remark The three masks in Fig.~\ref{fig:mask} are small regions cut from the high-resolution original mask. Therefore, the objects in these small masks can be considered as tiny targets. As shown in Fig.~\ref{fig:mask}, with the help of the range mask, HighlightNet is able to adjust the enhancement range at pixel level and realize the targeted enhancement of small objects.  Moreover, as shown in Fig.~\ref{fig:fig4}, HighlightNet can help the base tracker recognize small objects in low-light conditions.

\subsection{Transformer-Based Parameter Adjustment}
The main function of this module is producing constraint $\alpha$ and truncation threshold $\beta$ by processing global features. As mentioned in the last section, constraint $\alpha$ is fused with the enhanced range mask to obtain the actual enhancement of each pixel. Therefore, this branch has a similar influence on all pixels. Since global information plays a more important role than local context, Transformer~\cite{vaswani2017nips} is introduced. DETR~\cite{DETR} has shown the ideal efficiency of Transformer in processing global information in computer vision tasks. With the advantage of global processing, this module can dynamically regulate parameters to adapt to illumination variation caused by the high mobility of UAVs.

In this global processing module, the resolution of input does not play an important role. We find that the computation resource can be greatly saved without performance degradation by setting the resolution of input to $32 \times 32$. Therefore, as shown in Fig.~\ref{fig:main}, the input gray-scale image $\mathbf{G}_{i} \in \mathbb{R}^{1 \times H_{i} \times W_{i}}$ is first identically downsampled to a low-resolution image $\mathbf{L}_{i} \in \mathbb{R}^{1 \times 32 \times 32}$.  After processing by a convolutional layer, it is in the size of ${16 \times 16 \times 16}$ and then reshaped to ${16 \times (16 \times 16)}$ to enter the encoder of Transformer. Several identical layers make up the encoder and then two sublayers make up each layer. Each layer consists of two sub-layers. The first layer uses the concept of Multi-Head Attention \cite{dosovitskiy2020image} (MHA) to drive the model to concentrate on information from various positions and representation subspaces. MHA is defined as:

\begin{equation}
	\label{mha}
	{\mathrm{MHA}}( {\mathbf{Q}}, {\mathbf{K}}, {\mathbf{V}}) = {\mathbf{Cat}}( \mathrm{head_1}, \mathrm{head_2}, \cdots, \mathrm{head}_h){\mathbf{W}}^{\mathbf{O}}\ ,
\end{equation}
where ${\mathbf{Q}}, {\mathbf{K}}, {\mathbf{V}} \in \mathbb{R}^{n \times d_m}$ are input embedding matrices, $n$ is sequence length, $d_m$ is the embedding dimension, and $h$ is the number of heads. Each head is defined as:
\begin{equation}
	\label{att}
	\begin{split}
		\mathrm{head}_j = {\mathrm{softmax}} \left[\frac{\hat{\mathbf{Q}} \hat{\mathbf{K}}^{\rm T}}{\sqrt{{\mathbf{d_K}}}} \right]\hat{\mathbf{V}},\\
	\end{split}
\end{equation}
where $\mathbf{d_K}, \mathbf{d_V}$ are the hidden dimensions of the projection subspaces. $\hat{\mathbf{Q}},\hat{\mathbf{K}},\hat{\mathbf{V}}$ are query, key ,and value projected into the subspaces respectively, which can be defined as:

\begin{equation}
	\label{att1}		
		\hat{\mathbf{Q}}={\mathbf{QW}}_j^{\mathbf{Q}},
		\hat{\mathbf{K}}={\mathbf{KW}}_j^{\mathbf{K}},
		\hat{\mathbf{V}}={\mathbf{VW}}_j^{\mathbf{V}},	
\end{equation}
where ${\mathbf{W}}_j^{\mathbf{Q}},{\mathbf{W}}_j^{\mathbf{QK}} \in \mathbb{R}^{\mathbf{d_m} \times \mathbf{d_K}},{\mathbf{VW}}_j^{\mathbf{V}} \in \mathbb{R}^{\mathbf{d_m} \times \mathbf{d_V}}$ are learned matrices.

The second sub-layer is a fully connected feed-forward network (FFN). The connection between each sub-layer is a residual module followed by layer normalization. Therefore, the output of each sub-layer can be expressed as:
\begin{equation}\label{equ:transformer}
	\begin{split}
		\mathbf{F}'= \mathrm{LN}\left ( \mathrm{MHA}\left ( \mathbf{F} \right )+\mathbf{F} \right )\quad,\\
		\mathbf{F}''= \mathrm{LN}\left ( \mathrm{FFN}\left ( \mathbf{F}' \right )+\mathbf{F}' \right )\quad,\\
	\end{split}
\end{equation} 
where $\mathbf{F}'$ and $\mathbf{F}''$ denote the output of MHA and FFN, respectively. LN is layer normalization. The output of Transfomer is in the same size as input ${16 \times (16 \times 16)}$. In order to obtain two adaptive parameters, a simple fully connected layer(FC) followed by the sigmoid activation function is introduced to reshape the output to a bivector [$\alpha$,$\beta$]. Each dimensionality in this bivector is then translated to constraint $\alpha$ and truncation threshold $\beta$ by the parameter transform module, which is a simple linear conversion.

\subsection{Soft Truncation}
In this module, a soft truncation function $\mathbf{T}\left ( \mathbf{G} \right )$ is proposed to filter useless features brought by over-enhanced noise. The intensity of enhancement weakening is negatively correlated with the gray value. Moreover, considering the limitation of computational resources on the UAV platform, this function should be easy to compute. Therefore, a cubic function determined by two parameters is employed, which can be expressed as:
\begin{equation}\label{equ:truncation}
	\begin{split}
		\mathbf{T}_{i}\left (\textit{r,c}  \right )= -\tau \times \left ( \beta _{i}\, - \mathbf{G}_{i}\left (  \textit{r,c}\right )\right )^{3},\; \tau \beta _{i}^{3}\mathbf{T}_{i}\left (\textit{r,c}  \right )< 0 ,
		\\ 0\leqslant \textit{r}< H_{i} \, ,\,  0\leqslant \textit{c}< W_{i},
	\end{split}
\end{equation} 
where $\mathbf{G}_{i}\left ( \textit{r,c} \right )$ is the value of each pixel in gray-scale image $\mathbf{G}_{i} \in \mathbb{R}^{1 \times H_{i} \times W_{i}}$, $ \beta _{i}$ is the threshold of truncation, $\tau$ determines the reducing range. $\mathbf{T}_{i}\left (\textit{r,c}  \right )$ is the output of the truncation function with the codomain of $\left [\tau \beta _{i}^{3},0 \right )$. Therefore, the value of the output anti-noise mask is subtractive, which will apply to the pixel-wise addition module to reduce the enhancement range of dark area noise.

\Remark The threshold $ \beta _{i}$ of truncation is an adjustable parameter produced by the parameter adjustment module. Therefore, the result of soft truncation is not only based on the input grey-scale image but also influenced by the global features acting on Transformer. Threshold $ \beta _{i}$ increases in extreme low-light conditions to filter more useless features brought by noise.

For the same purpose, a non-reference loss function of dark area noise $\mathcal{L}_{\mathrm{dan}}$ is introduced. Since severe noise problems always appear in dark regions of the image, this loss function limits the enhancement range of shadows and night sky. The enhanced image is first divided into local regions in the size of ${16 \times16}$. The average enhancement range of pixels in the dark area of each region is then calculated. Therefore, $\mathcal{L}_{\mathrm{dan}}$ can be expressed as:
\begin{equation}\label{equ:lossdan}
	\begin{split}
		\mathcal{L}_{\mathrm{dan}}= \frac{1}{N}\sum_{i}^{N}\mathbf{A}_{i}\quad, 
	\end{split}
\end{equation}
where $N$ is the number of regions, $\mathbf{A}_{i} $ is the average enhancement range of pixels in the dark area of each region. The threshold to determine whether the pixel is in a dark area is set to 0.04. This threshold is half of the average of $ \beta _{i}$. Because the impact of soft truncation becomes obvious from this threshold. Except for the $\mathcal{L}_{dan}$, other three non-reference losses are employed to further improve the performance of HighlightNet. The total loss can be expressed as:
\begin{equation}\label{equ:loss}
	\begin{split}
		\mathcal{L}_{\mathrm{total}}= \lambda _{1}\mathcal{L}_{\mathrm{dan}}+\mathcal{L}_{\mathrm{spa}}+\lambda _{2}\mathcal{L}_{\mathrm{exp}}+\lambda _{3}\mathcal{L}_{\mathrm{tv}}\quad,
	\end{split}
\end{equation}
where $\mathcal{L}_{\mathrm{dan}}$ is dark area noise loss, $\mathcal{L}_{\mathrm{spa}}$ is spatial consistency loss, $\mathcal{L}_{\mathrm{exp}}$ is exposure control loss, $\mathcal{L}_{\mathrm{tv}}$ is the illumination smoothness loss, $\lambda_1$, $\lambda_2$, and $\lambda_3$ are the weights of losses. $\mathcal{L}_{\mathrm{spa}}$, $\mathcal{L}_{\mathrm{exp}}$, and $\mathcal{L}_{\mathrm{tv}}$ are first proposed in~\cite{Guo_2020_CVPR} as non-reference losses. Therefore, HighlightNet can be trained with unpaired data, which avoid the high cost of preparing a paired dataset for UAV tracking tasks.

\section{Experiment}
 To testify the effectiveness and robustness of HighlightNet for both online object selection and UAV object tracking, our experiments can be divided into two parts. For the template selection task, we quantitatively testify the performance of HighlightNet on the Part2 subset of SICE dataset~\cite{Cai2018TIP} and compare HighlightNet with SOTA low-light enhancers, respectively Darklighter~\cite{Ye2021IROS}, EnlightenGAN~\cite{EnlightenGAN}, Zero-DCE~\cite{Guo_2020_CVPR}, and LIME~\cite{LIME}. As a preprocessing step for UAV target tracking, its success rate and precision on UAVDark135 benchmark with baseline are compared with other SOTA low-light enhancers to testify its advantage on UAV nighttime tracking tasks. Performance on sequences with different attributes in UAVDark135 is tested to verify its ability to deal with specific UAV challenges. Moreover, to demonstrate its universality for different UAV trackers, it has been implemented on 4 SOTA trackers, \textit{i.e.}, SiamAPN++~\cite{Cao2021IROS}, SiamAPN~\cite{Fu_2021_TGRS}, HiFT~\cite{Cao2021ICCV}, and SiamRPN++~\cite{LiBo_2019_CVPR}. To show the function of each module, an ablation study is also introduced. Finally, we conduct the real-world test by adopting HightlightNet on a typical UAV platform to verify its applicability. 


\subsection{Implementation Details}
To bring the capability of the illumination-based adaptive parameter adjustment into full play, the unpaired training set includes both low-light and over-exposed images. Therefore, images from Part1 of the SCIE dataset~\cite{Cai2018TIP} are employed to train our network. The training images are resized to 512×512. The weights $\lambda_1$, $\lambda_2$, and $\lambda_3$ are set to 200, 50, and 20 respectively, to balance the scale of losses. We implement our framework with PyTorch on a PC with an Intel i9-9920X CPU, an NVIDIA TITAN RTX GPU, and 32GB RAM. The batch size is set to 8, with a total of 100 epochs. We employ the ADAM optimizer and set the learning rate to 0.001. Other parameters of the optimizer are default. Finally, to confirm the viability of HighlightNet in nighttime UAV tracking, the real-world test uses an NVIDIA Jetson AGX Xavier, which is widely applied on UAV platforms. 

\subsection{Evaluation Metrics}
To verify HighlightNet’s advantage on the nighttime online object selection task, we perform quantitative experiments on the Part2 subset of SCIE dataset~\cite{Cai2018TIP}, which includes 229 multi-exposure sequences. To testify its performance in low-light conditions, we choose the first low-light image in all 229 sequences and resize them to a size of 960×640×3. Finally, we obtain 229 paired low/normal light images. We choose the Part2 subset of the SCIE dataset for it contains numerous outdoor sequences and is primarily designed for low-light enhancement.

We also use visual tracking evaluation measures to rate the performance because HighlightNet is also targeting nighttime UAV tracking. The studies follow the one-pass evaluation method~\cite{Mueller2016ECCV}, which uses two metrics: precision and success rate. The precision is calculated by measuring the CLE between the estimated position and the ground truth position. And the success rate is calculated using the intersection over union (IoU) between the estimated bounding box and the ground truth.

\subsection{Efficacy of HighlightNet for Online Target Selection}
Since target selection is mainly executed by human operators, human perception should be evaluated quantificationally to testify the efficacy of HighlightNet. Therefore, the peak signal-to-noise ratio (PSNR), and structural similarity (SSIM) are employed. TABLE~\ref{tab:psnr} reports the comparison result of HighlightNet with other SOTA low-light enhancers. HighlightNet achieves the highest PSNR and SSIM score on the Part2 subset of the SICE dataset~\cite{Cai2018TIP}. Darklighter is specially designed for trackers and thus its performance is unsatisfying in the test for human perception. Since initial target selection is inevitable, HightlightNet is more practical in nighttime UAV applications.

\begin{table}[htbp]
	\centering
	\vspace*{-2pt}
	\caption{Comparison of HighlightNet with other SOTA enhancers in terms of full-reference image quality assessment metrics. The best result is in \NoOne{red} whereas the second best one is in \textcolor{green}{green}. HighlightNet has an obvious advantage on facilitate human perception.}
	\resizebox{0.95\linewidth}{!}{
		\begin{tabular}{cccccc}
			\toprule
			\multicolumn{1}{c}{\multirow{2}{*}{Method} }& \multicolumn{1}{c}{Darklighter} & \multicolumn{1}{c}{LIME} & \multicolumn{1}{c}{Zero-DCE} & \multicolumn{1}{c}{EnlightenGAN} & \multicolumn{1}{c}{\textbf{HighlightNet}} \\
			\multicolumn{1}{c}{ } &\multicolumn{1}{c}{~\cite{Ye2021IROS}}&\multicolumn{1}{c}{~\cite{LIME}}&\multicolumn{1}{c}{~\cite{Guo_2020_CVPR}}&\multicolumn{1}{c}{~\cite{EnlightenGAN}}&\multicolumn{1}{c}{\textbf{(ours)}} \\
			\midrule
			PSNR$\uparrow$  & 8.3     & \textcolor{green}{\textbf{11.4}} & 8.9   & 10.8  & \textcolor[rgb]{ 1,  0,  0}{\textbf{11.8}}  \\
			SSIM$\uparrow$  & 0.57    & 0.70 & 0.63  & \textcolor{green}{\textbf{0.72}}  & \textcolor[rgb]{ 1,  0,  0}{\textbf{0.75}}  \\
			
			\bottomrule
	\end{tabular}}%
    \vspace{-12pt}
	\label{tab:psnr}%
	
\end{table}%

\begin{figure*}[!t]	
	\centering
	
	\includegraphics[width=0.49\linewidth]{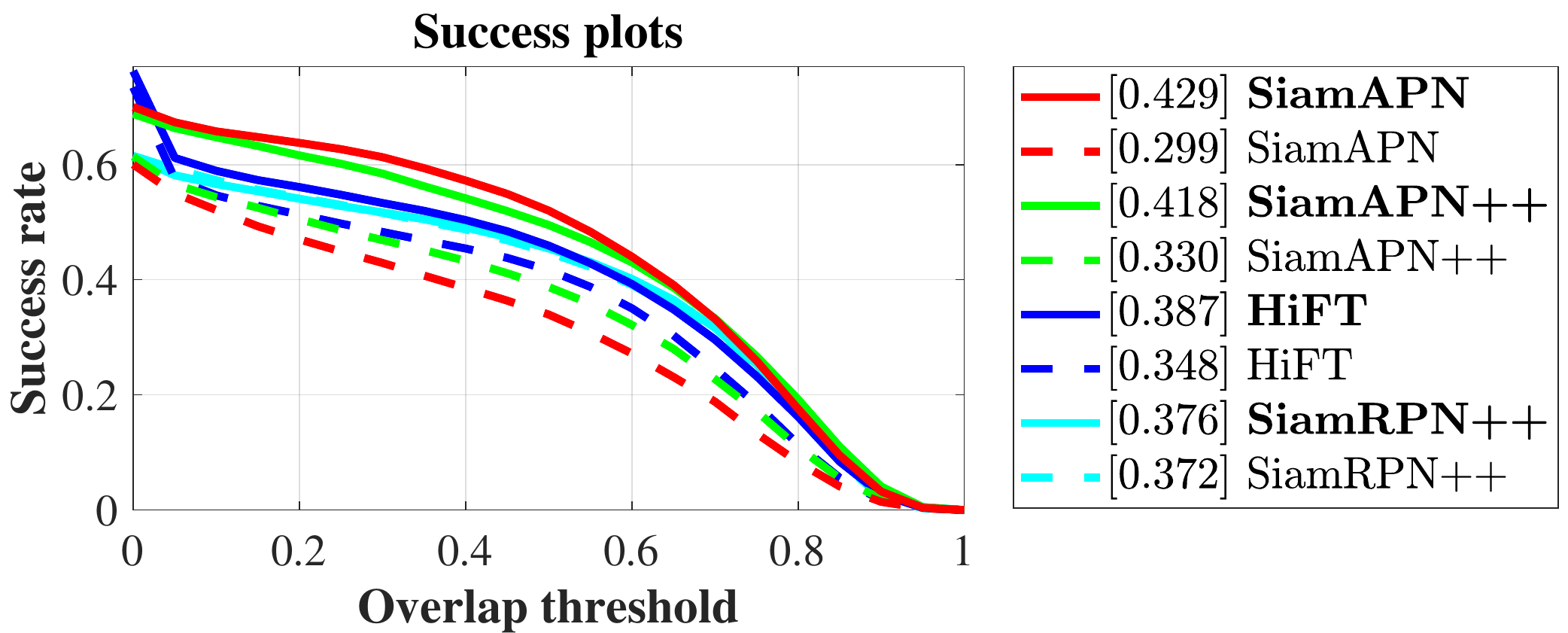}
	\includegraphics[width=0.49\linewidth]{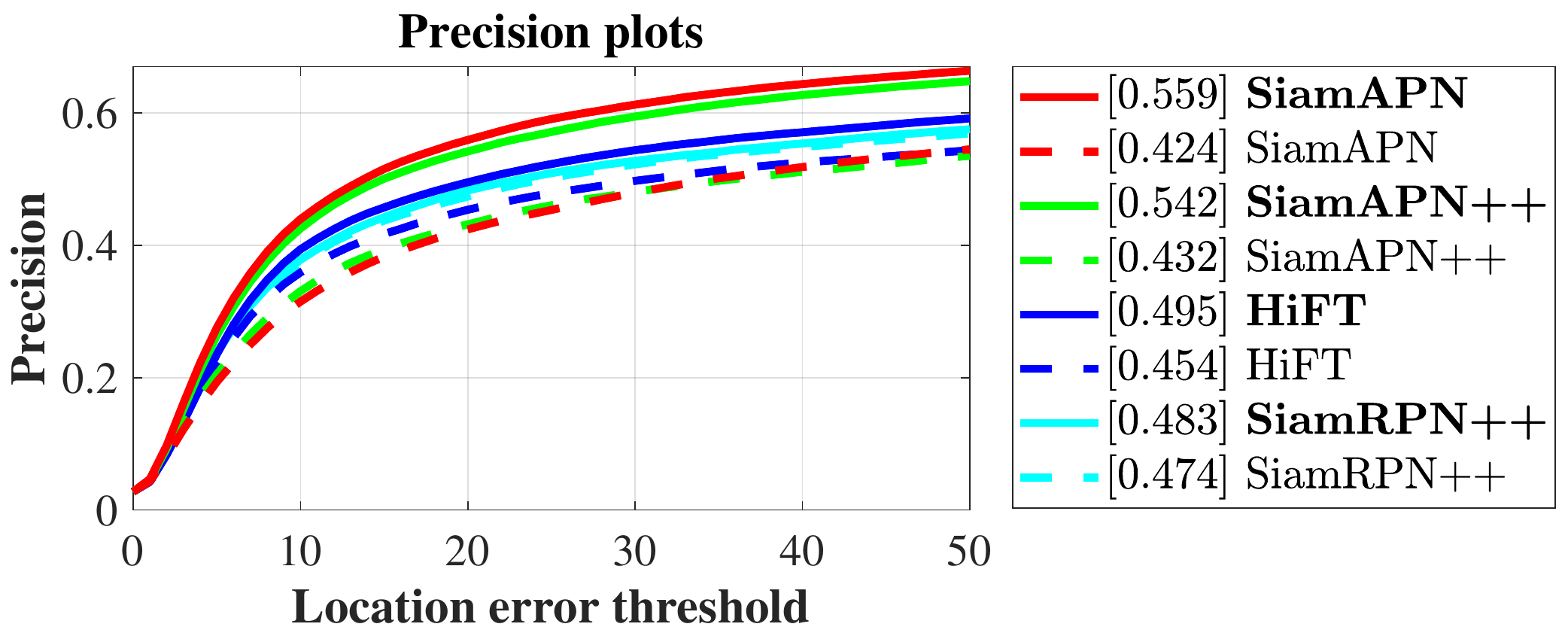}
	\setlength{\abovecaptionskip}{-4pt} 
	
	\caption
	{
		Precision and success plots of SOTA trackers with HighlightNet enabled or not. The results with HightlightNet are in \textbf{bold} type. HighlightNet improves the performance of all involved trackers.
	}
	\label{fig:all}
	\vspace{-0.4cm}
\end{figure*}

\Remark HighlightNet not only has a better performance on the object selection task but also be system-friendly in terms of computational complexity, for the design of HighlightNet takes the computational cost into account heavily.

\begin{table}[b!]
	\scriptsize
	\centering
	\vspace{-12pt}
	\caption{Comparison of HighlightNet with SOTA enhancers on UAVDark135. The first ,second, and third best results are in \NoOne{red}, \textcolor{green}{green}, and \textcolor{blue}{blue} font, respectively. $\Delta_s$ and $\Delta_p$ represent the percentages of success rate and precision exceeding the baseline, \textit{i.e.}, original tracker.}
	\begin{tabular}{ccccc}
		\toprule
		Method       & \multicolumn{1}{c}{\, Succ. \,}   & \multicolumn{1}{c}{\, Prec. \,}  &$\Delta_s$ (\%) & $\Delta_p$ (\%) \\ 
		\midrule
		Origin         & 0.336                         & 0.428                    & - & -   \\
		
		\midrule
		
		+LIME~\cite{LIME}         & 0.415                         & 0.518                   & 23.5 &   21.0 \\
		\midrule
		+Darklighter~\cite{Ye2021IROS}         & 0.415                         &  \textcolor{blue}{\textbf{0.529}}                   &23.5  &  \textcolor{blue}{\textbf{23.8}}  \\
		\midrule
		+EnlightenGAN~\cite{EnlightenGAN} & \textcolor{blue}{\textbf{0.418}}                      & 0.522  & \textcolor{blue}{\textbf{24.4}}  &   22.0 \\
		\midrule
		+Zero-DCE~\cite{Guo_2020_CVPR}     & \textcolor{green}{\textbf{0.419}}                         & \textcolor{green}{\textbf{0.530}} & \textcolor{green}{\textbf{24.7}}  &  \textcolor{green}{\textbf{23.8}}  \\
		\midrule
		{\textbf{+HighlightNet (ours)}}         &\textcolor[rgb]{ 1,  0,  0}{\textbf{0.424}} & \textcolor[rgb]{ 1,  0,  0}{\textbf{0.539}}  &\textcolor[rgb]{ 1,  0,  0}{\textbf{26.2}} & \textcolor[rgb]{ 1,  0,  0}{\textbf{25.9}}   \\
		\bottomrule
	\end{tabular}
    \vspace{-12pt}
	\label{tab:compare1}
\end{table}


\subsection{Efficacy of HighlightNet for UAV Dark Tracking}
\subsubsection{Comparison of HighlightNet and SOTA enhancers}
To demonstrate the superiority of HighlightNet for low-light UAV tracking, the performance of HighlightNet and other SOTA low-light enhancers---Darklighter~\cite{Ye2021IROS}, EnlightenGAN~\cite{EnlightenGAN}, Zero-DCE~\cite{Guo_2020_CVPR}, and LIME~\cite{LIME}, on UAVDark135 benchmark with tracker SiamAPN++~\cite{Cao2021IROS} is analyzed. SiamAPN++ is employed because it's designed specifically for UAV tracking. TABLE~\ref{tab:compare1} shows the tracking results with images enhanced by various enhancers. Since HighlightNet is mainly designed for UAV nighttime applications, it achieves the highest improvement in success rate and precision among the four enhancers. HighlightNet brings an increase of more than \textbf{26\%} and \textbf{25\%} in success rate and precision, surpassing the second-best Zero-DCE by \textbf{6.07\%} and \textbf{8.8\%} in the increase of success rate and precision, respectively.

\subsubsection{HighlightNet on different UAV challenges}
Sequences in benchmark UAVDark135 with different attributes are employed to testify the tracking performance of specific challenges in UAV applications. The improvement of tracking performance in typical low-light aerial challenges, \textit{i.e.}, fast motion (FM), illumination variation (IV),  low-resolution (LR), visual occlusion (OCC), and viewpoint change (VC) is reported in TABLE~\ref{tab:att}. As shown, in the term of IV and FM, it achieves an enormous improvement of over \textbf{25\%}, which is credited to the design of dynamic parameter adjustment according to global illumination. Furthermore, thanks to the pixel-level illumination enhancement brought by HightlightNet, the ability of trackers to cope with LR is also revived markedly. Though HighlightNet is not specially designed to deal with OCC and VC, it still brings an increase of over \textbf{13\%}, which is due in large part to the capacity of HighlightNet to highlight potential features for the tracker.

\begin{table}[b!]
	\scriptsize
	\centering
	\caption{Evaluation of tracking performance in different UAV challenges, results are shown as success/precision. $\Delta$ denotes the percentages of improvement brought by HighlightNet.  In all typical challenges of dark tracking, HighlightNet boosts tracking performance obviously.}
	\resizebox{0.95\linewidth}{!}{
		\begin{tabular}{cccccc}
			\toprule
			Attributes       & \multicolumn{1}{c}{ FM }   & \multicolumn{1}{c}{ IV }  & \multicolumn{1}{c}{LR }& \multicolumn{1}{c}{ OCC} & \multicolumn{1}{c}{VC} \\ 
			\midrule
			Original         & 0.315/0.404  &   0.312/0.408   & 0.323/0.464  & 0.332/0.416  & 0.353/0.426 \\
			\midrule
			+HighlightNet (ours)      & 0.399/0.507  &  0.398/0.515  & 0.402/0.571  &  0.392/0.497 & 0.408/0.485  \\
			\midrule
			$\Delta$ (\%)    &{\textbf{26.7/25.5}} &{\textbf{27.6/26.2}} &{\textbf{24.4/23.1}}  &{\textbf{18.1/19.5}} &{\textbf{15.6/13.8}} \\
			\bottomrule
	\end{tabular}}
	\label{tab:att}
\end{table}

\subsubsection{HighlightNet on different trackers}
As shown in Fig.~\ref{fig:all}, HighlightNet facilitates the low-light tracking performance significantly, with obvious improvement for all trackers in both precision and success rate. Among these four trackers, HighlightNet improves the dark tracking performance of SiamAPN and SiamAPN++ obviously, with an increase of more than \textbf{42.6\%} and \textbf{26.2\%} in success rate, as well as an increase of \textbf{31.8\%} and \textbf{25.9\%} in precision. 

\begin{figure}[!t]
	\centering	
	\includegraphics[width=0.99\linewidth]{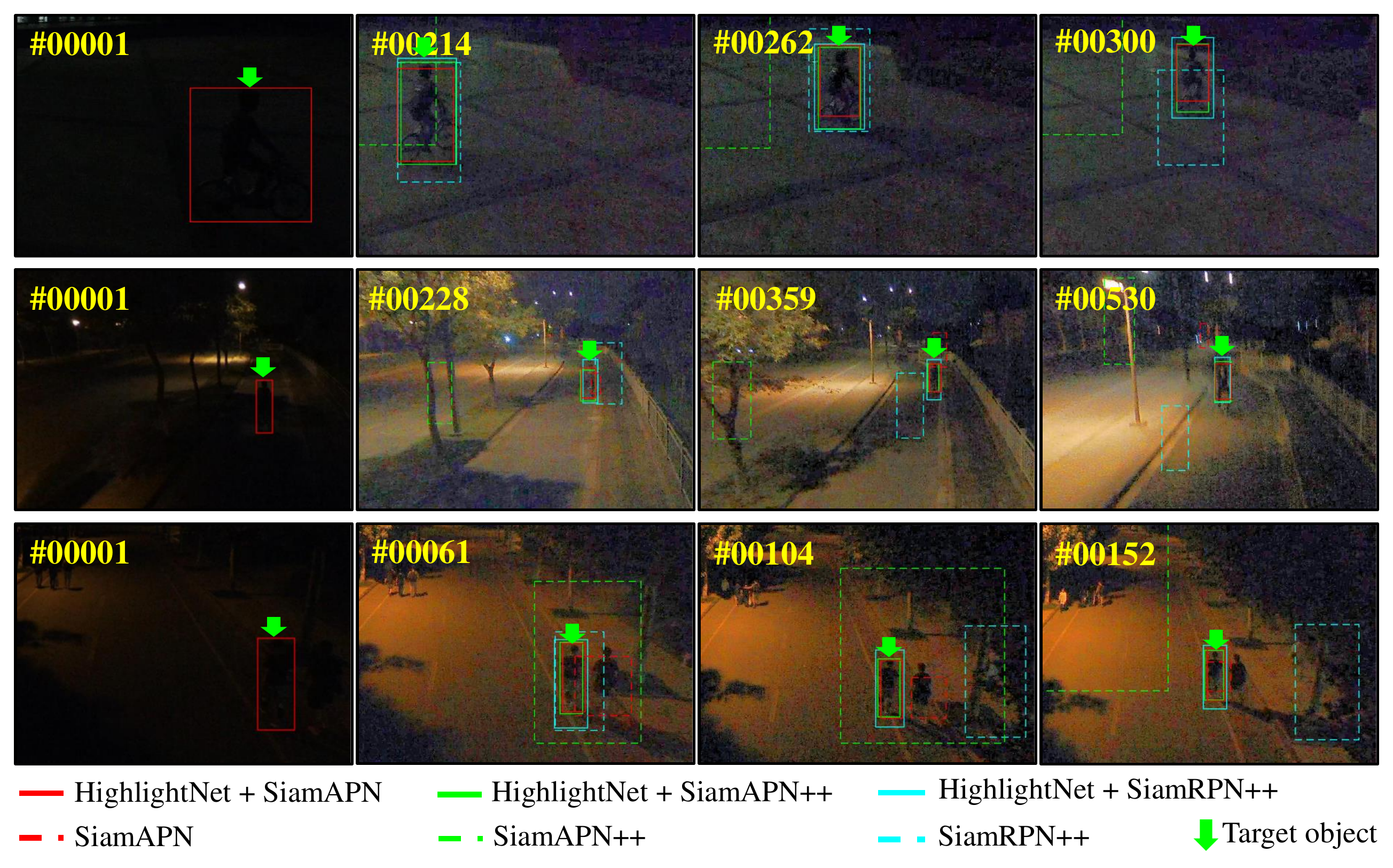}
	\setlength{\abovecaptionskip}{-4pt} 
	\caption
	{
		Qualitative results of trackers with HighlightNet enabled (solid boxes) or not (dashed boxes). For visualization, HighlightNet is employed to enhance the images in the last three columns. The sequences, from top to bottom, are $bike3$, $bike9$, and $group1$ from the public benchmark UAVDark135. HighlightNet dramatically improves tracker nighttime tracking performance.
	}
	\label{fig:qua}
	\vspace{-0.4cm}
\end{figure}

Some tracking screenshots of the trackers with or without HighlightNet are exhibited in Figure~\ref{fig:qua}. As shown, with the help of HighlightNet, trackers are able to retrieve the target missing in the shadow. We can conclude that HighlightNet boosts the accuracy and reliability of the trackers in these low-light conditions.

\Remark For visualization, images in Fig.~\ref{fig:qua} are enhanced by HighlightNet except for the first column. The enhanced results demonstrate that HighlightNet can markedly facilitate the perception of human operators in nighttime conditions, benefiting the initial annotation task.

\begin{table}[!b]
	\scriptsize
	\centering
	\caption{Comparison of HighlightNet with different modules enabled. Success rate and precision (Succ./Prec.) represent tracking performance while PSNR and SSIM represent human perception. The best results are in \NoOne{red} font.}
		\begin{tabular}{cccccc}
			\toprule
			\multicolumn{1}{c}{RM} & \multicolumn{1}{c}{TPA} & \multicolumn{1}{c}{ST} & \multicolumn{1}{c}{$\mathcal{L}_{\mathrm{dan}}$} & Succ./Prec. & PSNR/SSIM \\
			\midrule
			&        &        &        & 0.336/0.428 & -/- \\
			\midrule
			\multicolumn{1}{c}{\checkmark} & \multicolumn{1}{c}{\checkmark} &        &        & 0.391/0.501 & 11.6/0.67 \\
			\multicolumn{1}{c}{\checkmark} & \multicolumn{1}{c}{\checkmark} & \multicolumn{1}{c}{\checkmark} &        & 0.419/0.530 & 11.3/0.73 \\
			\multicolumn{1}{c}{\checkmark} & \multicolumn{1}{c}{\checkmark} &        & \multicolumn{1}{c}{\checkmark} & 0.416/0.527 & 11.6/0.69 \\
			\midrule
			\multicolumn{1}{c}{\checkmark} &        & \multicolumn{1}{c}{\checkmark} & \multicolumn{1}{c}{\checkmark} & 0.419/0.526 & 10.6/0.73 \\
			\midrule
			& \multicolumn{1}{c}{\checkmark}& \multicolumn{1}{c}{\checkmark} & \multicolumn{1}{c}{\checkmark} & 0.418/0.524 & 11.2/0.71 \\
			\midrule
			\multicolumn{1}{c}{\checkmark} & \multicolumn{1}{c}{\checkmark} & \multicolumn{1}{c}{\checkmark} & \multicolumn{1}{c}{\checkmark} & \textcolor[rgb]{ 1,  0,  0}{\textbf{0.424/0.539}} & \textcolor[rgb]{ 1,  0,  0}{\textbf{11.8/0.75}} \\
			\bottomrule
		\end{tabular}
		\label{tab:abla}%
	\end{table}%

\subsection{Ablation Study}
The performance of different variants of HighlightNet is investigated in this subsection. Tracking results on UAVDark135~\cite{li2021allday} are exhibited in TABLE~\ref{tab:abla}. RM, TPA, and ST denote the range mask, the Transformer-based parameter adjustment, and the soft truncation, respectively. We trained all these variations in the same way. The bottom row shows the unsatisfying performance without HighlightNet, while the top row shows the completely enhanced result. Since the soft truncation and $\mathcal{L}_{\mathrm{dan}}$ both target filtering of useless features, their effects are discussed in the same sector.

\subsubsection{Range mask} To ablate the range mask without impacting other modules, it is replaced with another mask. Each pixel in this mask is the same as the average value of pixels in the range mask. As shown in the sixth line of TABLE~\ref{tab:abla}, without the range mask, the improvement of the tracking performance is reduced to \textbf{8.0\%} in success rate and \textbf{8.5\%} in precision. Moreover, with a decrease of \textbf{5.0\%} and \textbf{6.0\%}, PSNR and SSIM also degrade obviously. The function of the range mask to make HighlightNet focus on the tracking target and improve illumination uniformity can be verified.
\subsubsection{TPA} Constraint $\alpha$ and threshold $\beta$ are set to constant values, which are their average values in a complete HighlightNet. Since the parameters are invariable, the adjustment branch is consequently disabled. As shown in the fifth line of TABLE~\ref{tab:abla}, the gains of HighlightNet bringing to both human perception and tracking performance degraded significantly. Therefore, the Transformer-based parameter adjustment module is of clear benefit to UAV tracking tasks.
\subsubsection{ST \& $\mathcal{L}_{\mathrm{dan}}$} Ablating ST and $\mathcal{L}_{\mathrm{dan}}$ respectively, the benefit of introducing HighlightNet decreases to some extent, validating the effectiveness of both anti-noise mask and dark area noise loss. Furthermore, as shown in the second line of TABLE~\ref{tab:abla}, simultaneously disabling ST and $\mathcal{L}_{\mathrm{dan}}$ brings a more obvious reduction in success rate, precision, PSNR, and SSIM. Therefore, it is testified that ST should be employed along with the $\mathcal{L}_{\mathrm{dan}}$ to minimize the over-enhancement of noise.

\begin{figure}[!t]

	\centering	
	\includegraphics[width=0.99\linewidth]{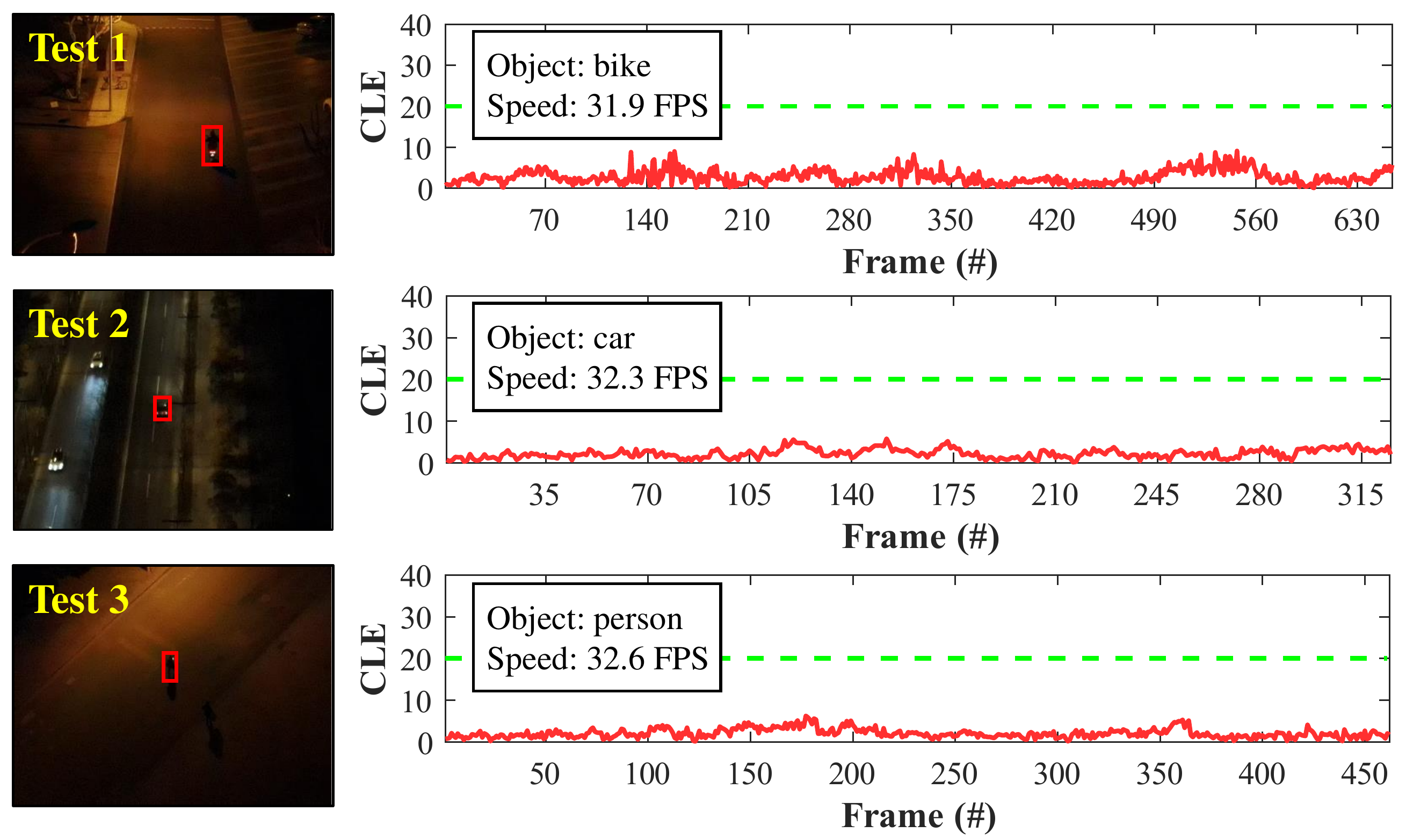}
	\setlength{\abovecaptionskip}{-4pt} 
	\caption
	{
		Evaluations on a standard UAV platform in the real world. The estimated positions are indicated by \NoOne{red} bounding boxes. Below are CLE curves between predictions and ground truth. The \NoTwo{green} dashed line marks a 20-pixel threshold, within which tracking mistakes are usually considered acceptable. The base tracker achieves good nighttime tracking with the help of HighlightNet.
	}
	\label{fig:real}
	\vspace{-0.4cm}
\end{figure}

\subsection{Real-World Tests}

To testify HighlightNet's practicability in real-world low-light UAV tracking applications, we employ it on a typical UAV platform with embedded devices, \textit{i.e.}, the NVIDIA Jetson AGX Xavier. HighlightNet offers a conspicuous real-time performance of $\textbf{32.2}$ FPS even without TensorRT acceleration. In addition, Fig.~\ref{fig:real} shows CLE curves and various real-world nighttime tracking experiments. The main obstacles in the testing are illumination variation, low brightness, tiny target, and partial occlusion. The prediction errors are less than 20 pixels on the CLE curves, showing that the tracking is accurate. With the help of HighlightNet, the basic tracker can provide reliable object tracking in nighttime conditions.

\section{Conclusion}
This work proposes a low-light enhancer HightlightNet for facilitating both online target selection step and tracking step in UAV nighttime tracking. Therefore, HighlightNet takes both human perception and nighttime challenges for UAVs into consideration. By three novel modules, it highlights the potential features and consequently reduces the influence of rapid illumination variation, artificial light sources, small objects, and image noise. Experiments on various tracking approaches confirm its compatibility and effectiveness in both online target selection and tracking. Comparison with other SOTA low-light enhancers verifies the advantages of HighlightNet for facilitating human perception and tracking performance in nighttime conditions. Real-world experiments on a typical UAV platform confirm its applicability and dependability while consuming little computational resources. To summarize, we are confident that this research will aid in the expansion of UAV tracking applications to nighttime environments.

\section*{Acknowledgment}

This work is supported by the National Natural Science Foundation of China (No. 62173249), the Natural Science Foundation of Shanghai (No. 20ZR1460100).


\bibliographystyle{IEEEtran}
\normalem
\bibliography{IROS2021}

\end{document}